\documentclass[10pt,twocolumn,letterpaper]{article}

\usepackage{cvpr}
\usepackage{times}
\usepackage{epsfig}
\usepackage{graphicx}
\usepackage{amsmath}
\usepackage{amssymb}
\usepackage[x11names]{xcolor}
\usepackage{etoolbox}
\patchcmd{\thebibliography}{\section*{\refname}}{}{}{}

\newif\ifproofread

\usepackage[breaklinks=true,bookmarks=false]{hyperref}

\cvprfinalcopy 

\def\cvprPaperID{****} 

\begin{document}

\title{The Effect of Learning Strategy versus Inherent Architecture Properties on the Ability of Convolutional Neural Networks to Develop Transformation Invariance}

\author{Megha Srivastava\\
Stanford\\
Computer Science Department\\
{\tt\small meghas@stanford.edu}
\and Kalanit Grill-Spector\\Stanford\\Psychology Department\\
{\tt\small kgs@stanford.edu}
}

\maketitle
\proofreadtrue
\begin{abstract}
As object recognition becomes an increasingly common task in machine learning, recent research demonstrating neural networks’ vulnerability to attacks and small image perturbations show a need to fully understand the foundations of object recognition in order to build more robust models. Our research focuses on understanding the mechanisms behind how neural networks generalize to spatial transformations of complex objects. While humans excel at discriminating between objects shown at new positions, orientations, and scales, past results demonstrate that this may be limited to familiar objects -  humans demonstrate low tolerance of spatial-variances for purposefully constructed novel objects. Because training artificial neural networks from scratch is similar to showing novel objects to humans, we seek to understand the factors influencing the  tolerance of artificial neural networks to spatial transformations. 

We conduct a thorough empirical examination of seven  Convolutional Neural Network (CNN) architectures. By training on a controlled face image dataset, we measure changes in accuracy of the model across different degrees of 5  transformations: position, size, rotation, resolution transformation due to Gaussian blur, and resolution transformation due to resample. We also examine how learning strategy affects generalizability by examining the effect different amounts of pre-training have on model robustness. Overall, we find that the most significant contributor to transformation invariance is pre-training on a large, diverse image dataset. Moreover, while AlexNet tends to be the least robust network, VGG and ResNet architectures demonstrate higher robustness for different transformations. Along with kernel visualizations and qualitative analyses, we examine  differences between learning strategy and inherent architectural properties in contributing to invariance of  transformations, providing valuable information towards understanding how to achieve greater  robustness to transformations in CNNs.

\end{abstract}

\section{Introduction}

Biologically-inspired networks, such as deep convolutional neural networks (CNNs), have demonstrated success in many image classification and recognition tasks. However, recent works such as the development of adversarial inputs that fool the networks have demonstrated weaknesses in CNNs, necessitating the need to better measure and improve the robustness and generalizability of artificial neural networks. Although researchers have shown similarities between  the architectures and learned features of CNNs with certain properties of human and primates’ visual systems, it is not clear to what degree the biologically-inspired artificial neural networks derive their success from large training datasets versus intrinsic network properties. 

We focus on how  neural networks for object recognition and discrimination perform when seeing complex objects that have undergone spatial transformations. The human visual system is typically adept at this task independent of spatial transformation - we can recognize and discriminate complex objects that are presented at previously unseen positions, orientations, and resolutions. 

However,  CNNs' robustness to spatial transformations is  not well understood. The shift-equivariance properties of the convolution layers, together with the dimensionality reduction of pooling layers removing spatial information, provide these networks invariance to small amounts of shifts. However, other architecture properties such as depth and learning strategies of these networks may also affect the tolerance of CNNs to spatial transformations when performing object recognition. Because data collected for computer vision tasks may not always span different spatial variations, it is important to understand how learning strategy and inherent architecture properties affect CNN spatial transformation invariance. 

Our goal is to empirically study the role of learning strategy and CNN architecture properties on the performance of CNNs in complex object recognition over previously unseen spatial transformations. Specifically, the input to our models consists of images of human faces - a representative complex object. We study CNN architectures of varying depth and complexity, including recent networks that are commonly used in the Computer Vision community, and examine how the ability to generalize recognition of human faces to new positions, size, rotations, and resolutions is affected by the CNN architecture and learning strategy.

Our approach consists of (i) selecting a subset of CNN model architectures to implement, (ii) use \emph{transfer learning}~\cite{TLTutorial} to realize various learning strategies ranging from training the network on a set of faces from scratch to fine-tuning a pre-trained network, and (iii) study CNN performance when generalizing to new positions, sizes, rotation, and resolutions of faces not seen previously. With our results, we demonstrate how different CNN models vary in their ability to learn spatial transform generalizability as an intrinsic architecture property versus training exposure and learning strategy.

\section{Related Work}

Our research provides a comprehensive and empirical examination of spatial invariance properties of CNNs. Although there exist recent works on improving CNN transformation robustness, none have thoroughly examined the degree to which CNNs are capable of generalizing to different types of spatial transformations, nor comment on generalizability gained from intrinsic network property versus training exposure and learning strategy.  Broadly speaking, past research can divided into 1.~Small-scale investigations on transformation invariance over large, standard computer vision datasets containing images with varied transformations, and 2.~The development of CNNs with improved spatially transformation generalizability. These works complement recent psychology research on human spatial generalizability to novel objects. 

The closest work to our research is by Bunne, et. al.~\cite{bunne2018studying}, which analyzed how two pre-trained CNN architectures, AlexNet~\cite{krizhevsky2012imagenet} and ResNet~\cite{he2016deep}, behave under a series of 9 transformations by studying the softmax outputs of a network for specific classes. They found that the two networks were able to learn a small degree of invariance, but large changes in transformations, such as a large rotation of  picture of a broom, resulted in higher \emph{softmax} outputs for images such as a brush. However, the authors use pre-trained networks trained on large datasets such as images from the ImageNet Large Scale Visual Recognition Challenge, which contain millions of images containing multiple objects in a variety of positions and sizes. This results in decreased clarity on whether the source of any invariance is simply due to pre-training, versus an actual CNN model architecture property. For example, is the higher likelihood of a brush over a broom after significant rotation simply due to ImageNet containing more images of brushes that are rotated than brooms in different rotations? Therefore, for our work, we focus on understanding the effect of different degrees of pre-training, including training the different models from scratch. 

Similarly, Jaderberg et. al.~\cite{jaderberg2015spatial} introduced the Spatial Transformer, which provides existing CNNs the capability to spatially transform feature maps over individual data items. The module allows input images to be transformed into canonical class examples, on which prediction is then run across. However, no claim is made regarding the success of Spatial Transformation Network when the dataset itself does not contain images at a variety of spatial transformations, as they train over an MNIST dataset containing digits in a variety of positions, sizes, and rotations. They demonstrate success over distorted versions of MNIST datasets, but do not measure how much exposure during training time to new transformations is required. Our work, which seeks to empirically determine the amount of transformation perturbation required for a significant drop in performance, can possibly help inform the amount of transformation within input data needed for generalizability. Furthermore, our approach can help us understand the degree at which pre-processing methods such as random cropping aid in spatial transformation generalizability. 

Other computational research in this area includes, 
Lenc et. al.~\cite{lenc2015understanding}, which examined equivarience properties in AlexNet and found that deeper layer representations were more tuned to specific transformations, and Kauderer-Abrams et.al.~\cite{kauderer2017quantifying}, which determined data augmentation had the most significant effect on translation invariance for small models up to only four layers. Recent research has proposed separating transformation values from object representation, so that deep networks learn invariant features separately.  For example, Cohen et. al.~\cite{cohen2014transformation} proposed the use of G-Convolutions that exploit symmetry to achieve spatial invariance, Cohen et. al.~\cite{cohen2016group}  used group representation theory to represent objects independent of spatial pose, Anselmi et. al.~\cite{anselmi2016unsupervised} proved the ability to create invariant signatures for image patches in classification, and Hinton et. al., proposed the use of capsules to more learn high-information vector outputs that more efficiently enable spatial invariance than the scalar-value inputs of high-layer neurons in CNNs. Sabour et. al.~\cite{sabour2017dynamic} built upon the success of these capsule networks for invariance by proposing a dynamic routing strategy between capsules, while Shen et. al.~\cite{shen2017patch} proposed the use of patch before feeding features to the next layer in a CNN as a way to learn location invariance. Finally, both Simonyan et. al. and Szegedy et. al.~\cite{szegedy2016rethinking} focused on very deep networks, with more than 20 layers, and how the added computational complexity aids in learning invariant properties. 

Interestingly, recent psychology research has demonstrated that humans have difficulty generalizing to new spatial transformations when recognizing foreign objects that hold no social or semantic significance. Remus et. al.~\cite{Remus} found that while exposing foreign objects at a single position is not enough for humans to develop position-invariance when discriminating between objects, showing two positions improves generalizability significantly. We believe there are many useful links between understanding how object recognition and spatial invariance develops in both the  human visual system and artificial networks, and therefore want to similarly examine the thresholds for  generalizability to different spatial transforms in CNNs. 

Thus, none of the described past research provides a thorough understanding of the difference between training strategy and network properties in developing invariance, and the precise degree of tolerance towards spatial transformation variations the network has. By running experiments on networks trained from scratch, we better understand the effect of not only model architecture, but also training strategy. In the effort to properly understand how object recognition is learned, it is important to understand the exact degree different architectures can tolerate transformations and what training strategy is most effective. 

\section{Methods}

Using a dataset of human faces under different spatial transformations (translation, size, resolution, rotation, and pose), our research studies how the performance of convolutional neural networks at generalizing to new spatial transformations that were not present among the faces in the training set is affected by the structural complexity of the network and the training method. We approach our task with the following steps, which we proceed to describe in more details later in this section:
\begin{enumerate}
\item Vary CNN structural complexity by implementing 7 widely-used CNN architectures: {AlexNet, ResNet18, ResNet50, SqueezeNet, VGG11, VGG19, and AlexNet with a Spatial Transformer module}. 
\item Utilize transfer learning on a large, standard image dataset like ImageNet by adopting three different learning strategies: {training network from scratch with random initializations, training network after initializing with pre-trained model, and pre-trained model with last layer re-trained from scratch and all other layers frozen}
\item Train on a fixed {position/size/rotation/resolution} and, after hyperparameter tuning, measure network accuracy on images at parametrically increasing shifts of  {position/size/rotation/resolution}, thereby examining generalizability. For each experiment, only one transformation among {position/size/rotation/resolution} changes between training and testing. 
\item Run experiments across our 7 model implementations, 3 training strategies, and 5 different transformations (two different resolutions), resulting in a total of 90 experiment runs
\end{enumerate}
We detail our method and choices for each of the above four steps below. 

We chose 7 CNN architectures to implement, which vary not only in depth but also in architecture design. Our goal was to choose networks that are popularly used, as the impact of our work is most relevant to those using state-of-the-art networks who seek to understand the causes of spatial transformation invariance in their vision task. We first chose AlexNet~\cite{krizhevsky2012imagenet}, a popular 8-layer network with solid performance on the ImageNet Challenge. We then chose two variants of the VGG network, VGG-11 and VGG-1, which differ from AlexNet by being deeper, allowing us to understand whether architecture depth impacts spatial transformation invariance~\cite{simonyan2014very}. We furthermore chose two implementations of ResNet, which for several years have achieved state of the art accuracy metrics, and thus important to examine. The goal of ResNet is for the CNN to learn a residual value by skipping connections across layers, thus making deeper networks to perform better~\cite{he2016deep}.  We seek to determine whether the skipped connections allow the network to be more invariant to spatial transformations. Finally, we choose to implement SqueezeNet, which achieves high accuracy comparably to AlexNet with significantly fewer parameters, in order to understand whether the decrease in number of parameters allows the remaining network parameters to develop a greater amount of spatial invariance~\cite{iandola2016squeezenet}. 

Next, we developed three different learning strategies to adopt in our training: 
\begin{enumerate}
\item training from scratch with all layers randomly initialized, 
\item train after initialization with pre-trained model, and 
\item pre-trained model with all but the last layer frozen and the last layer trained from scratch after random initialization.

\end{enumerate}     All pre-training occured over an ImageNet data subset, which contains a large amount of natural image scenes in a variety of spatial transformations, unlike out fixed face dataset as described below. The three different learning strategies allow us to understand the necessity of pre-training on a large image dataset with natural scene images containing a variety of spatial transformations to achieve spatial invariance. By including a network trained from scratch, we differ from past research because we seek to answer whether spatial invariance can be achieved even when the training data consists of only one fixed transformation value.

Given a chosen CNN implementation and the three different learning strategies, we measure network accuracy and generalizability to one of 5 transformations: position (or translation) invariance, size (or scale) invariance, rotation invariance, resolution invariance from Gaussian blur, and resolution invariance from resampling. First, we train our model on fixed values of the 5 transformations. Then, given our fixed dataset of 101 faces described in the dataset section below, we generate modified images at fixed, staggered degrees of transformations for each of the 5 transformations. For translation invariance, we generate new positions up to a maximum of 140 pixel shift in any direction, defined by a circle with radius equal to the amount of shift, with steps of 10 pixels. At each step of 10 pixels shift up to 140 pixels, we measure the network accuracy on these new positions given the original trained model. Therefore, we are able to visualize how increase in position shift by pixels affects network accuracy, providing comprehensive information on translation invariance of the different models. 

For conducting size invariance experiments, we generated new sizes from scales of 0.4 to 2.4 of the given original size, allowing us to measure invariance for both increased and decreased sizes.  For rotation invariance, we generated new rotations of the faces by rotating them by steps of 15 degrees, from 15 degrees to 345 degrees. Lastly, for the two resolution invariances, we generated new images by either applying a Gaussian blur with 5 kernel amounts, or resampling at 5 different amounts. 

Finally, we run our experiments to measure generalizability for each of the 5 possible transformations across models and training types. During training the set of images, with fixed transformation values (i.e. center position, original size, no decrease in resolution), for each of the 101 faces is randomly divided in a 80:20 ratio with 12 images being put into training set and 3 into validation set. Therefore, there is class balance for all faces in the training and validation  set. The validation set is then replicated for the different transformations for the corresponding images that we previously generated. This ensures that the only difference between the different validation sets for a given transformation is the degree of transformation, as the face images and orientations are consistent across the validation set.  

During training, we did not adopt any cross-validation since out set of faces is fairly uniform and we ensured a balanced training and validation split on a per-face basis. Furthermore, we aimed to determine good values of hyper-parameters, including Adam vs. SGD with momentum optimization, learning rate, momentum, step size, gamma, and batch size, and discuss results of the hyper-parameter tuning in the Experiments section. After training, we plot how accuracy is affected by staggered shifts of all 5 different transformations, and analyze these plots in the Experiments section. Finally, we visualize the network kernels at the early convolutional layers in order to determine whether transformation invariance might be due to the presence of “edge detectors” as lower layer kernel features, as suggested about the human visual system \cite{hubel}.

\section{Dataset and Features}

The specific data set we used was collected in our lab, the Vision and Perception Neuroscience Laboratory in Stanford’s Psychology Department. With a fixed camera configuration and gray background, photos were taken of 101 subject with faces rotated at 15 degree increments from -105 to +105 along the vertical axis. The subjects used were college-aged students at Stanford University, and each face, regardless of rotation, is centered with the image frame and surrounded by a gray background. All images are in grayscale with original size 2272 px by 1704 px, and resolution 72 pixels/inch.  Depending on the network we use, we crop the images to fit the input image size - in all cases, the original image is larger than the input size and therefore no loss of image information is incurred. A sample from our dataset is shown in the figure below:

\begin{figure}[!htbp]
\begin{center}
\includegraphics[width=0.8\linewidth]{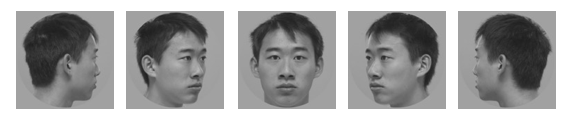}
\end{center}
\caption{Images of faces at different orientations corresponding to the same subject}\label{fig:datasetfaces}
\end{figure}

In order to measure invariance towards different spatial transformations, we perturb the images to achieve the desired transformations as described in the previous section. All transformations of our data are made with standard Python image libraries.

\section{Experiments, Results and Discussion}

\subsection{Model Training with Hyper-parameter Tuning}\label{Exp:MT}

The first step in our experiments involved training the various models for later use in evaluating the impact of spatial transformations. As noted earlier, we considered seven network models (AlexNet~\cite{krizhevsky2012imagenet}, SqueezeNet 1.1~\cite{iandola2016squeezenet}, VGGNet-11, VGGNet-19~\cite{simonyan2014very}, ResNet-18, ResNet-50~\cite{he2016deep}, and AlexNet with a Spatial Transformer module~\cite{jaderberg2015spatial,STNTutorial} which we refer to as {\it AlexSTNet}. We trained models for various architecture on images of faces using our dataset which consisted of 15 poses for each of the 101 faces with poses ranging from -105 degree to 105 degree in steps of 15 degrees. 

For all but the AlexSTNet architecture we learnt three different models depending on the level of pre-training on ImageNet. In each case the final fully connected layer was replaced with a new randomly initialized one with 101 outputs, and the three learnt models differed in how the other layers were handled. In Experiment 1, all the other layers were pre-initialized on ImageNet but then finetuned on the face imaget. In Experiment 2, all the other layers were pre-trained on ImageNet but then their weights were frozen and thus unaffected by training on face data set.  In Experiment 3, there was no pre-training and new weights were learnt starting from random initialization. For AlexSTNet we were unable to successfully pre-train the network on ImageNet and so only conducted an Experiment 4 which like Experiment 3 involves no pre-training.  

During training we aimed to find good values of following hyper-parameters: optimization algorithm (SGD with momentum vs Adam), learning rate, momentum (in case of SGD), step size, gamma, and batch size. To do so, we plotted the progression of both accuracy and loss vs. epoch number, and modified the hyperparameters based on visual assessment of the shape and gap between the curves for training vs. validation data. 

\begin{figure*}[t]
\begin{center}
\fbox{
\includegraphics[width=0.32\linewidth]{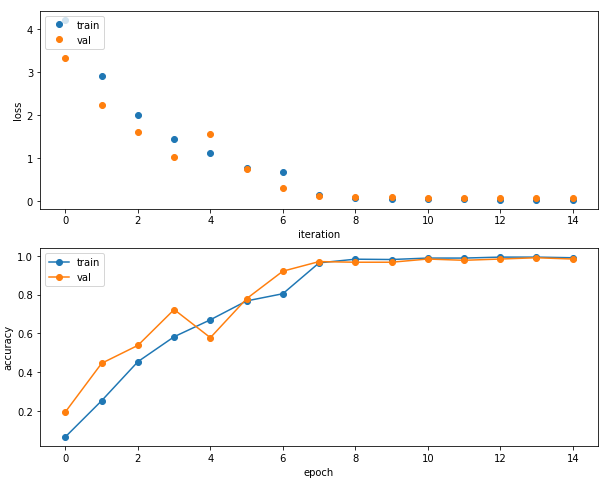}
\includegraphics[width=0.32\linewidth]{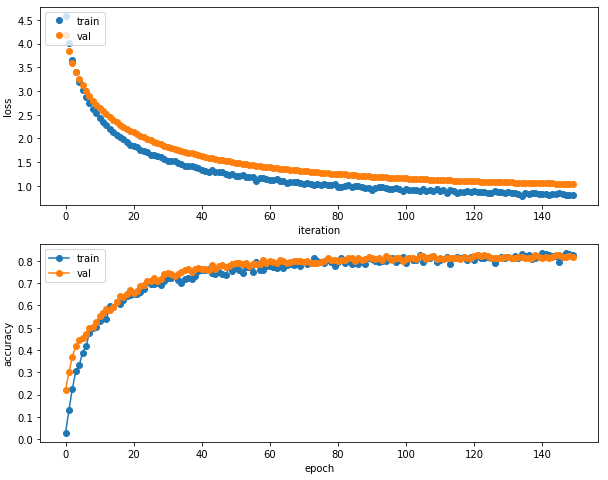}
\includegraphics[width=0.32\linewidth]{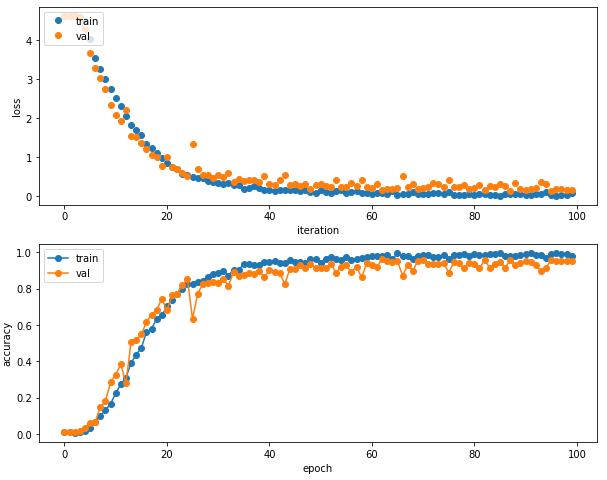}
}
\end{center}
\caption{Loss and Accuracy vs. Epoch \# plots for VGGNet-11 under Experiments 1 (pre-trained layers finetuned), 2 (pre-trained layers frozen), and 3 (training from scratch, i.e. no pre-training). Figures for other network models are under the Appendices}\label{fig:vn11}
\end{figure*}

Table~\ref{tab:hp} shows the values of hyperparameters that were found to yield good results, while Figure~\ref{fig:vn11} show the resulting accuracy and loss vs. epoch number curves for VGGNet-11. For reasons of space we have shown here the curves only for VGGNet-11 but all the plots are included in Figures~\ref{apfig:astn} through \ref{apfig:rn50} in the Appendices. Generally we see that in almost all cases the  training resulted in generally good validation accuracy, in most cases above 90\%.

\begin{table}
\begin{center}
\begin{tabular}{|l|c|l|}
\hline
Expt. \# & Model & Final Hyperparameters \\
\hline\hline
1 & All & SGD, lr=0.001, momentum=0.1,\\
  &     & step\_size=7, gamma=0.1,\\
  &     & batch\_size=4\\
\hline
2 & AlexNet & Adam, lr=0.000075,\\
  & SqueezeNet  & step\_size=10, gamma=0.9,\\
    &     & batch\_size=8\\
\hline
2 & ResNet-18 & Adam, lr=0.0001,\\
  & ResNet-50  & step\_size=10, gamma=0.9,\\
  & VGGNet-11 & batch\_size=8\\
\hline
2 & VGGNet-19 & Adam, lr=0.0002,\\
  &   & step\_size=10, gamma=0.9,\\
  &           & batch\_size=8\\
\hline
3 & AlexNet & Adam, lr=0.0001,\\
  & SqueezeNet & step\_size=20, gamma=0.9,\\
    &     & batch\_size=8\\
\hline
3 & ResNet-18 & Adam, lr=0.00001,\\
  & ResNet-50 & step\_size=20, gamma=0.9,\\
  & VGGNet-19 & batch\_size=8\\
\hline
3 & VGGNet-11 & Adam, lr=0.0001,\\
  &   & step\_size=20, gamma=0.9,\\
  &   & batch\_size=8\\
\hline
4 & AlexNet & Adam, lr=0.00003,\\
  & with Spatial& step\_size=20, gamma=0.9,\\
    & Transformer & batch\_size=8\\
\hline
\end{tabular}
\end{center}
\caption{Hyperparameters found for different experiments and network architectures.}\label{tab:hp}
\end{table}

\subsection{Testing with Diverse Spatial Transforms}

The second part of our experimental work focused on studying how the models trained as described in the previous subsection perform when tested against images of the faces that have been subjected to various spatial transformation.  Recall, our models were trained using face images only at one position, size, resolution, and rotation. Note also that the face images  in our dataset are not present in ImageNet, and the set of labels (corresponding to the identity of faces) has no correspondence to ImageNet labels. So the experiments described here show the invariance to various spatial transformations on complex objects that were previously not seen except under one spatial transformation.  

We studied the following five spatial transformations, and the metric we used was accuracy which is an appropriate metric considering the goal of this namely, namely the invariance exhibited by the models to spatial transformations. 

\subsubsection{Translation}
\begin{figure*}[!htbp]
\begin{center}
\fbox{
\includegraphics[width=0.32\linewidth]{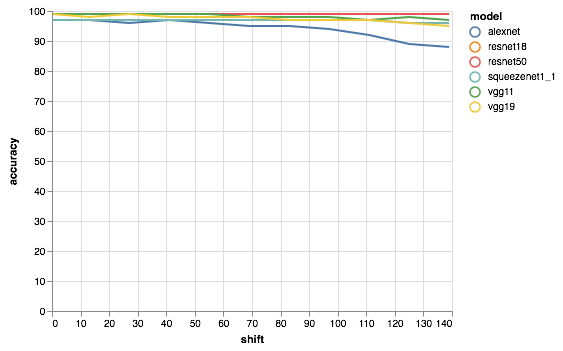}
\includegraphics[width=0.32\linewidth]{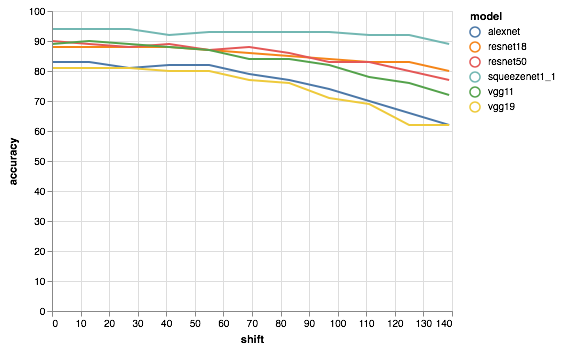}
\includegraphics[width=0.32\linewidth]{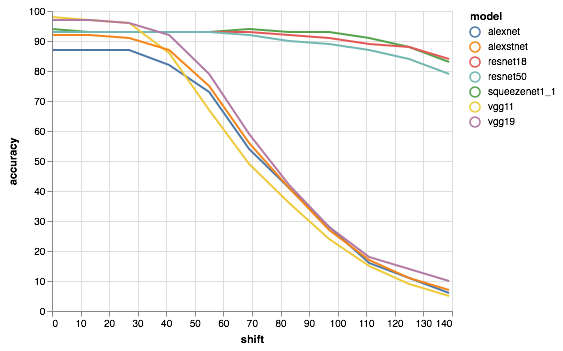}
}
\end{center}
\caption{Performance of different convnets with training images at the center and test images at varying amount of translation under Experiments 1, 2 and 3.}
\label{fig:translate}
\end{figure*}

Here we subjected the original images to translations in random directions with the amount of shift selected randonly between 0 and the maximum possible such that the face still stayed within the original image boundaries. Form~\ref{fig:translate} we make several key observations. Firstly, models trained under Experiment 1, where layers pre-trained on ImageNet were finetuned, did the best while those from Experiments 2 (pre-trained layers were frozen) and Experiment 3 (no pre-training) did increasingly worse. Secondly, accuracy falls with amount of translation. Thirdly, in Experiment 3, SqueezeNet, ResNet-18, and ResNet-50 performed much better than AlexNet, VGGNet-11, VGGNet-18, and even AlexNet with the Spatial Transformation module~\cite{jaderberg2015spatial}. These observations lead to the following conclusions. Firstly, preetraining on ImageNet helped quite a lot in attaining translation invariance, suggesting that exposure to objects under different translations as in ImageNet heloped. Second, certain network architectures, specifically SqueezeNet, ResNet-18, and ResNet-50, have structural properties that give them translation invariance to larger amounts of shift. Note that one would expect all convolution networks to exhibit invariances to small amounts of shift due to the combined effects of convolutional layers and pooling. It is also surprising that the AlexNet with a Spatial Transformation module did not perform as well, suggesting that Spatial Transformation module does not provide any intrinsic invariance due to its structure and its primary advantage may be that it learns invariances faster or more easily as suggested in ~\cite{jaderberg2015spatial}.

\subsubsection{Resizing}

\begin{figure*}[!htbp]
\begin{center}
\fbox{
\includegraphics[width=0.32\linewidth]{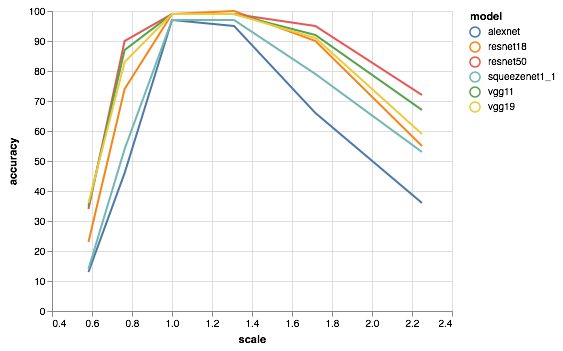}
\includegraphics[width=0.32\linewidth]{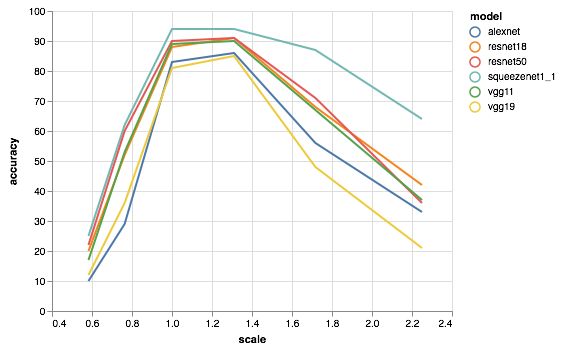}
\includegraphics[width=0.32\linewidth]{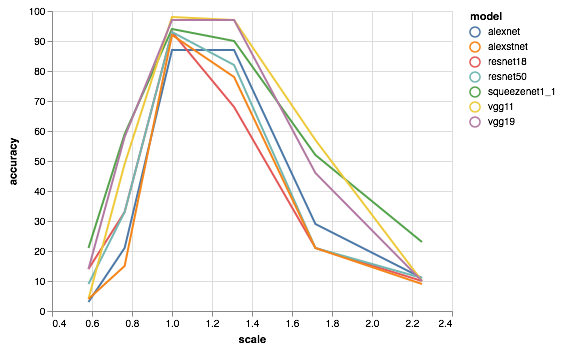}
}
\end{center}
\caption{Performance of different convnets with training images at the center and test images at varying amount of scaling under Experiments 1, 2 and 3.}
\label{fig:resize}
\end{figure*}

Here we resized the faces while keeping them centered in the image, resizing them both up and down by various amounts. Specifically we scaled them by factors of 2.25 (which is the maximum possible), 1.717, 1.31, 1, 0.763, and 0.582. As seen in~\ref{fig:resize}, the more the face is resized, the worse off is the accuracy, and this occurs irrespective or pre-traing. So unlike translation, pre-training does not appear to help that much, and there are no systematic trends due to network architecture. However, a noteworthy observation is the asymmetry, i.e. there is a steeper degradation when scaling down by a factor than when scaling up by the same factor, possibly because of the loss of resolution that happens when image is made smaller.

\subsubsection{Rotation}

\begin{figure*}[!htbpp]
\begin{center}
\fbox{
\includegraphics[width=0.32\linewidth]{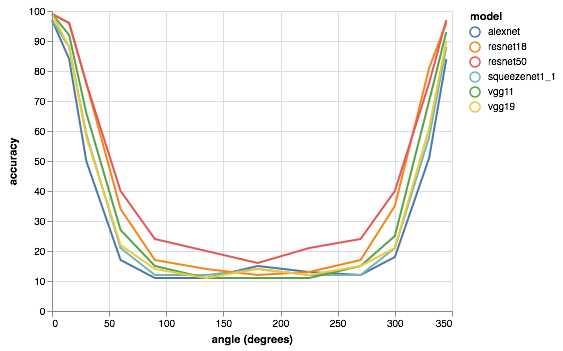}
\includegraphics[width=0.32\linewidth]{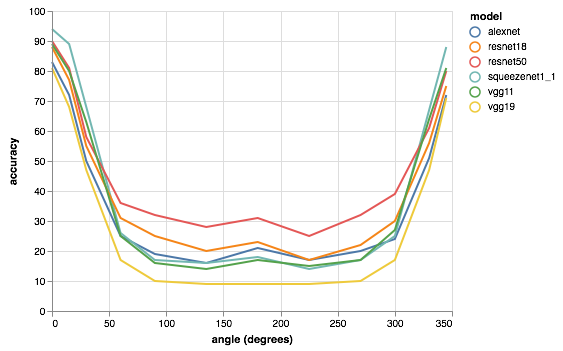}
\includegraphics[width=0.32\linewidth]{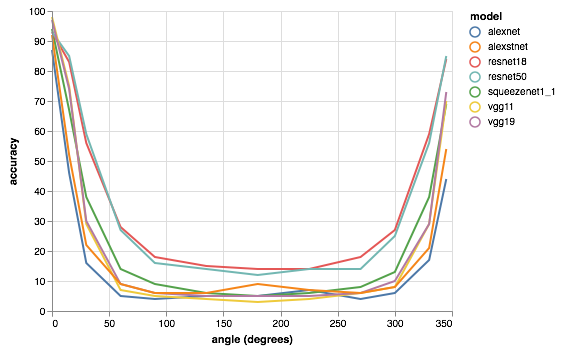}
}
\end{center}
\caption{Performance of different convnets with training images at the center and test images at varying amount of rotation under Experiments 1, 2 and 3.}
\label{fig:rotation}
\end{figure*}

Here we rotated the faces to various angles in the 0 to 360 degree range while keeping them centered in the image. Specifically we scaled them by 15, 30, 60, 90, 135, 180, 225, 270, 300, 330, and 345 degrees counter-clockwise. While the general trend of larger deviation from the original resulting in worse performance holds here as well~\ref{fig:rotation}, two observations stand out. Firstly, after a steep fall off in the first 50 degrees or so of rotation on either side, the performance then plateaus. Secondly, the two ResNets seem to perform distinctly better than the other architectures in the case when all layers are trained from scratch, suggesting that they have some degree of structural advantage for rotational invariance. 

\subsubsection{Resolution Reduction}

\begin{figure*}[!htbp]
\begin{center}
\fbox{
\includegraphics[width=0.32\linewidth]{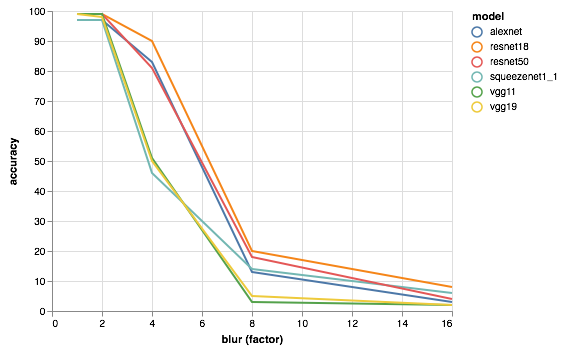}
\includegraphics[width=0.32\linewidth]{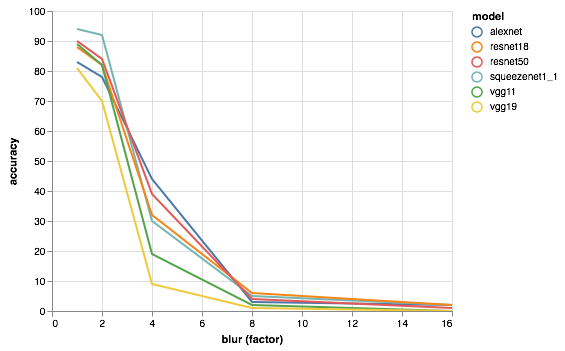}
\includegraphics[width=0.32\linewidth]{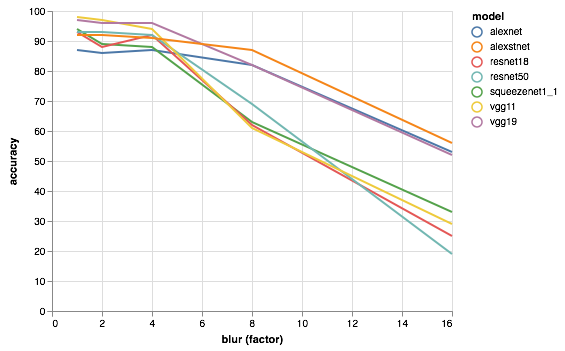}
}
\end{center}
\caption{Performance of different convnets with training images at the center and test images at varying amount of resolution under Experiments 1, 2 and 3.}
\label{fig:resample}
\end{figure*}

In this test we reduced the resolution of the original images by varying amount by discarding pixels. This was accomplished by rescaling the faces to smaller sizes and then scaling them back to the original size, in the process causing loss of resolution. Specifically we reduced resolution by factors of 2, 4, 8, and 16. Interestingly, as seen in~\ref{fig:resample} the performance was distinctly better in Experiment 3 as compared to Experiments 1 and 2, suggesting that pre-training om ImageNet hurt invariance to this transformation. Moreover in Experiment 3, AlexNet, AlexNet with Spatial Transformation module, and VGG-19 performed significantly better that the others.

\subsubsection{Gaussian Smoothing}

\begin{figure*}[!htbp]
\begin{center}
\fbox{
\includegraphics[width=0.32\linewidth]{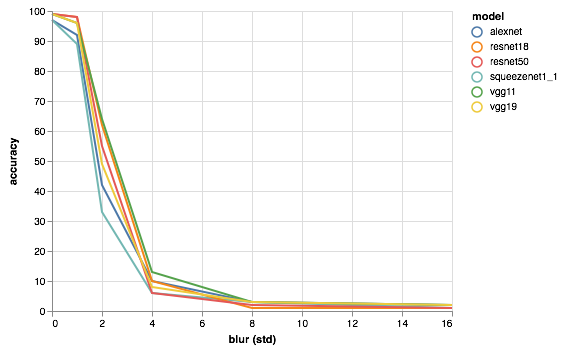}
\includegraphics[width=0.32\linewidth]{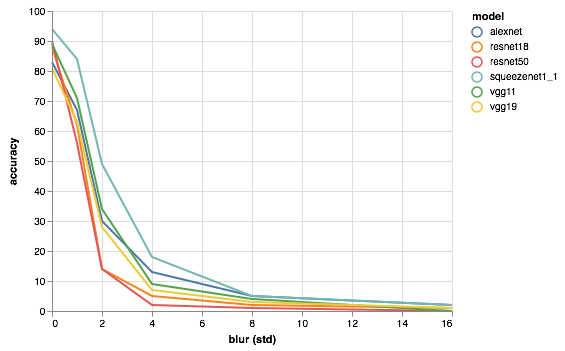}
\includegraphics[width=0.32\linewidth]{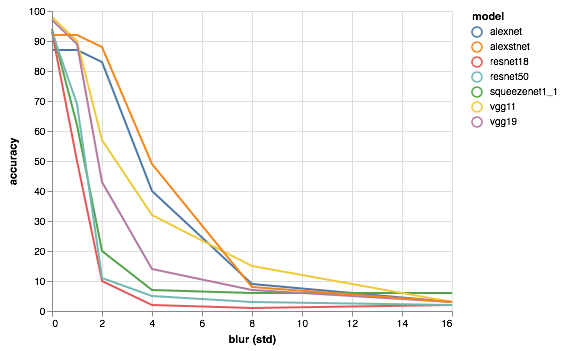}
}
\end{center}
\caption{Performance of different convnets with training images at the center and test images at varying amount of gaussian smoothing under Experiments 1, 2 and 3.}
\label{fig:gaussian}
\end{figure*}

In this test we subjected the image to Gaussuan smoothing with  standard deviations of 2, 4, 8, and 16.  As seen in~\ref{fig:gaussian}, the performance degrades with smoothing in all cases, but in Experiment 3 AlexNet, AlexNet with Spatial Transformation module, VGG-11, and VGG-19 did distinctly better than the others, which is a trend we saw in Resolution Reduction as well above. These observation further suggests that pre-training hurt invariance to resolution loss, and that AlexNet and VGG family of architectures appears to have some strucutral advantage. 

\subsection{Visualization}

Finally, we examined kernel visualizations in order to see if there is a more interpretable understanding to explain the differences between learning strategies' effects on robustness to transformation. Due to time constraints, we were unable to examine techniques to visualize more complex layers, such as those in SqueezeNet or the higher layers of AlexNet, and so we present visualizations of the 1st Convolutional Layer Kernels in AlexNet. Figure 7 visualizes the kernels when the layers pre-trained on ImageNet are fine-tuned, Figure 8 visualizes the kernels when the layer pre-trained on ImageNet are frozen, and finally Figure 9 visualizes the kernels when the entire network is trained from random initialization. 

Our visualizations show that in both pre-training strategies, the first layer  kernels show strong edges at different orientations,  corresponding to V1 layer neurons in the human visual system. The high similarity between Figures 7 and 8 suggest that fine-tuning the pre-trained layers does not effect the 1st-layer kernels, and weight changes must be occurring at higher layers. However, Figure 9 shows kernels with a high degree of noise, with faint orientations perceived only in some of the kernels. These visualizations suggest that unless the network is exposed to a dataset of objects in a variety of spatial transformations, the network is not incentivized to learn kernels such as "edge detectors" over more local features specific to the fixed transformation value of the training set. This highlights the possible dependence of spatial transformation invariance on 1st layer features of artificial networks. 

\begin{figure}[!htbp]
\begin{center}
\includegraphics[width=0.8\linewidth]{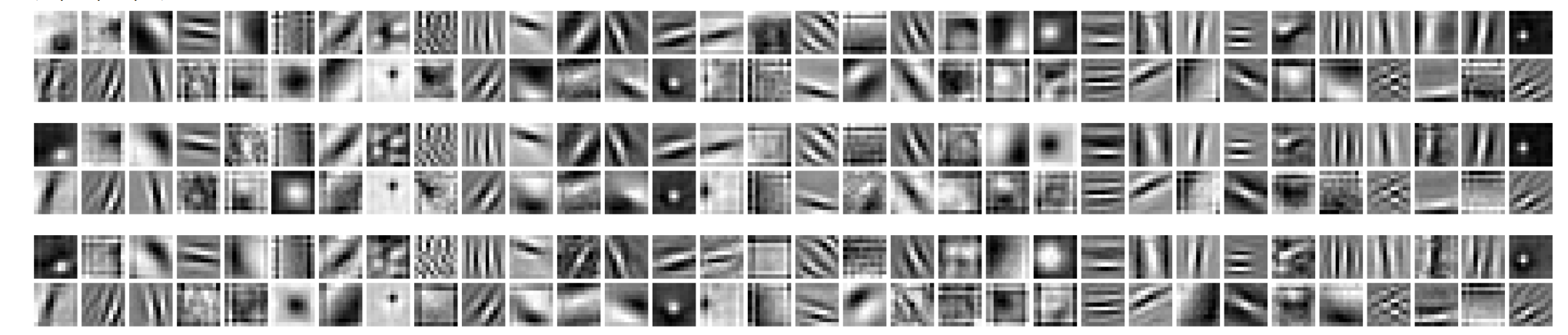}
\end{center}
\caption{ Visualization of 1st Convolutional Layer Kernels for AlexNet when Fine-Tuning pre-trained Layers }\label{fig:kernel_ftalex}
\end{figure}

\begin{figure}[!htbp]
\begin{center}
\includegraphics[width=0.8\linewidth]{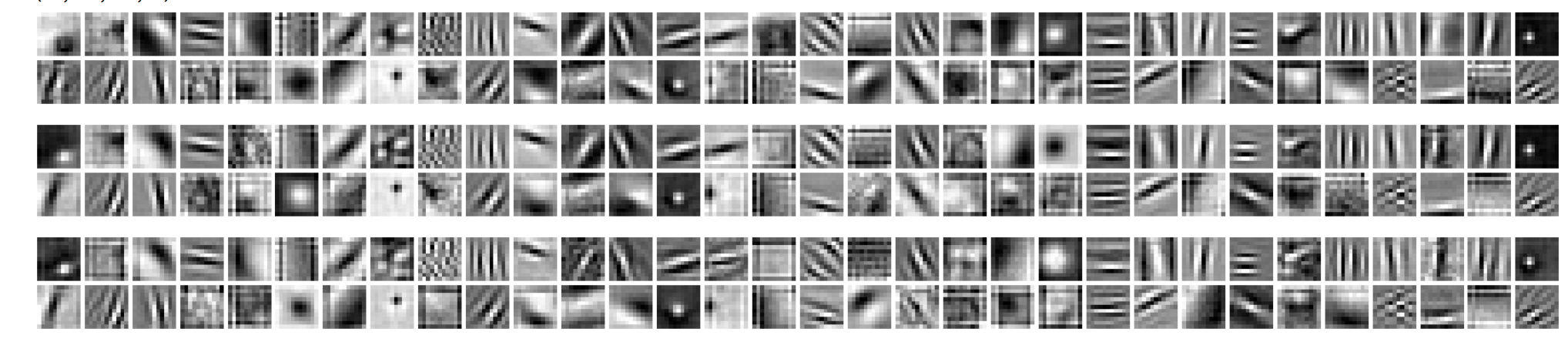}
\end{center}
\caption{Visualization of 1st Convolutional Layer Kernels for AlexNet when  pre-trained Layers are Frozen}\label{fig:kernel_ffalex}
\end{figure}

\begin{figure}[!htbp]
\begin{center}
\includegraphics[width=0.8\linewidth]{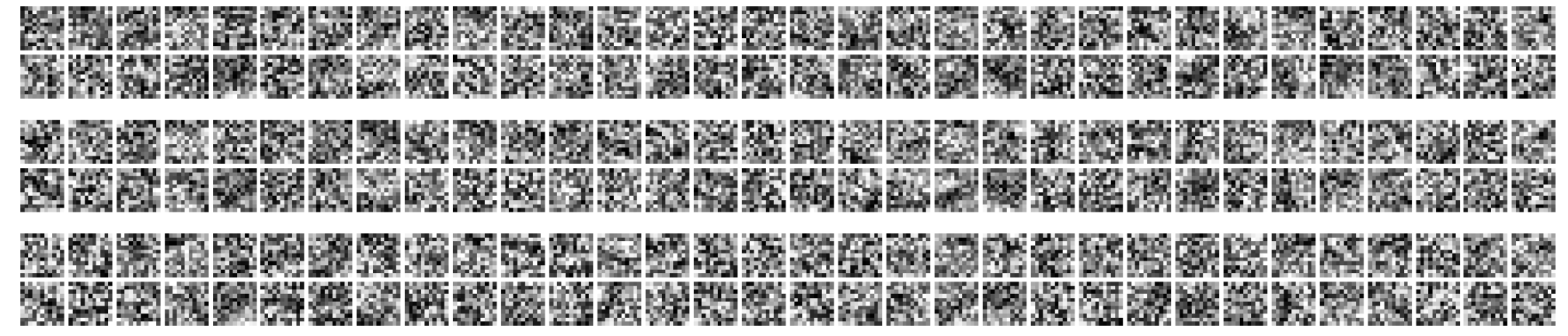}
\end{center}
\caption{Visualization of 1st Convolutional Layer Kernels for AlexNet when all layers trained from random initialization}\label{fig:kernel_newalex}
\end{figure}

\section{Conclusions}

Overall, our results demonstrate several interesting trends described earlier relating to the impact of pre-training, which in some cases appears to be beneficial (e.g.  invariance to translation) but in others cases detrimental (e.g. invariance to resolution), as well as the differences across the various architectures in their intrinsic invariance to various spatial transformation (e.g. our observation that ResNet performs better than others under rotation).

Although we attempted to implement the Spatial Transformer Network module for architectures other than AlexNet, we ran into software issues and were unable to find  implementations to build upon,  suggesting a high degree of complexity for the  design. However, as discussed previously, it is not clear whether Spatial Transformer Networks still rely on pre-training, and understanding whether the amount of pre-training needed is reduced is future work. 

Furthermore, human experimental results demonstrate that  exposure to only one fixed position is not sufficient for transformation invariance on a foreign object \cite{Remus}. This agrees with our conclusion of the importance of pre-training - which we can liken to visual experience or evolution for humans. However, the results do show that two or more positions are sufficient in improving human visual invariance. We  ran preliminary experiments in which the models are trained on two positions and sizes rather than one, but robustness in the non pre-trained setting was still low. More  analysis of this multiple transformation value exposure strategy will be  useful in drawing a link between artificial neural network and human visual generalizabiltiy. 

As CNNs become increasingly widespread in society, it is important to understand whether they solve object recognition tasks in a similar way as humans do, and whether a deeper understanding of the human visual system can improve biologically-inspired neural networks' robustness. We examine spatial invariances, an important property in how we visually understand our world, and demonstrate the importance of consider the effect of training strategy, and not just inherent architecture properties, on the invariance abilities of different commonly-used CNN architectures.

\clearpage
\section{Appendices}

\subsection{Loss and Accuracy vs. Epoch \# Plots for Different Network Architectures and Experiments}

\begin{figure}[h]
\begin{center}
\includegraphics[width=0.7\linewidth]{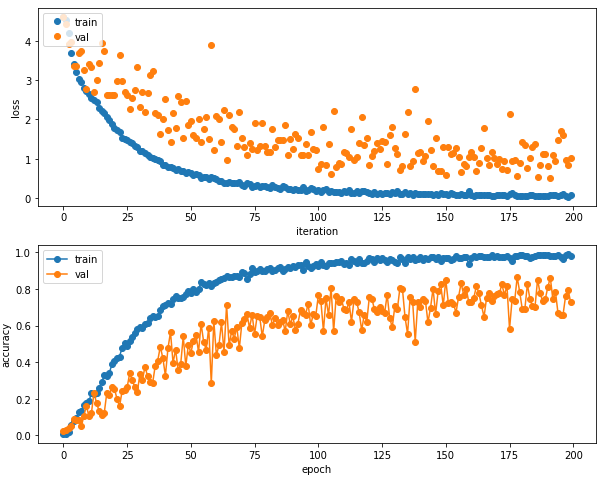}
\end{center}
\caption{Loss and Accuracy vs. Epoch \# plots for AlexNet with Spatial Transformation module while training from scratch.}\label{apfig:astn}
\end{figure}

\begin{figure*}[t]
\begin{center}
\fbox{
\includegraphics[width=0.32\linewidth]{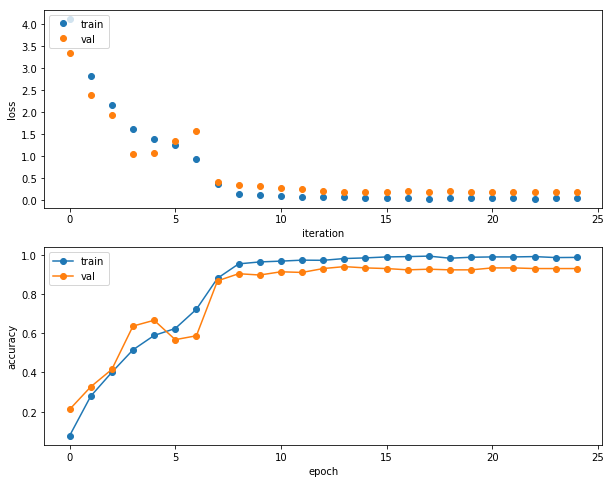}
\includegraphics[width=0.32\linewidth]{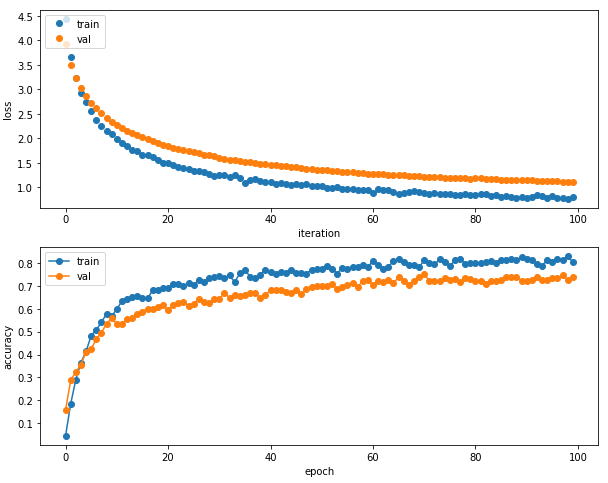}
\includegraphics[width=0.32\linewidth]{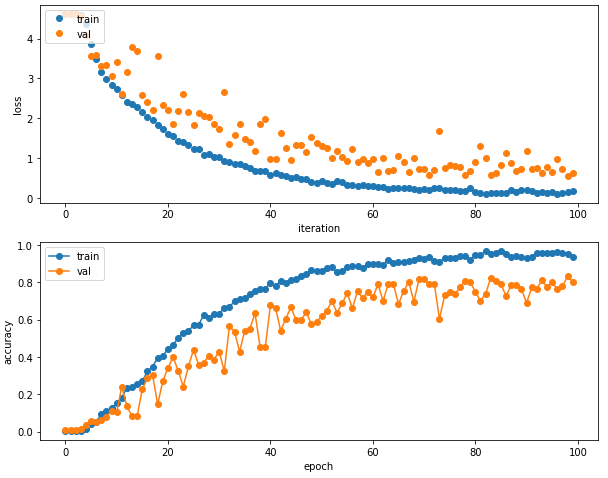}
}
\end{center}
\caption{Loss and Accuracy vs. Epoch \# plots for AlexNet under Experiments 1 (pre-trained layers finetuned), 2 (pre-trained layers frozen), and 3 (training from scratch, i.e. no pre-training).}\label{apfig:an}
\end{figure*}

\begin{figure*}[t]
\begin{center}
\fbox{
\includegraphics[width=0.32\linewidth]{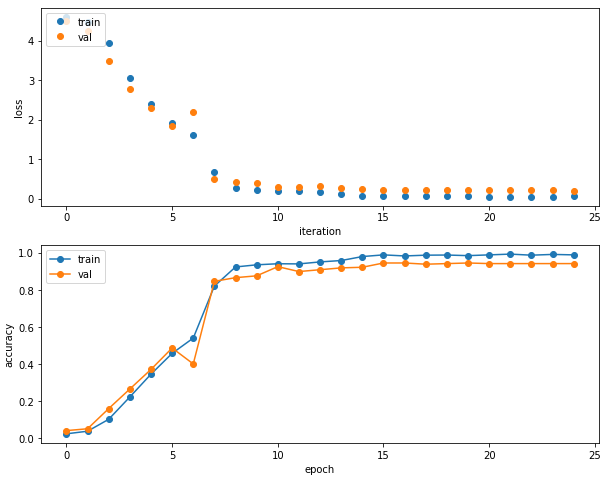}
\includegraphics[width=0.32\linewidth]{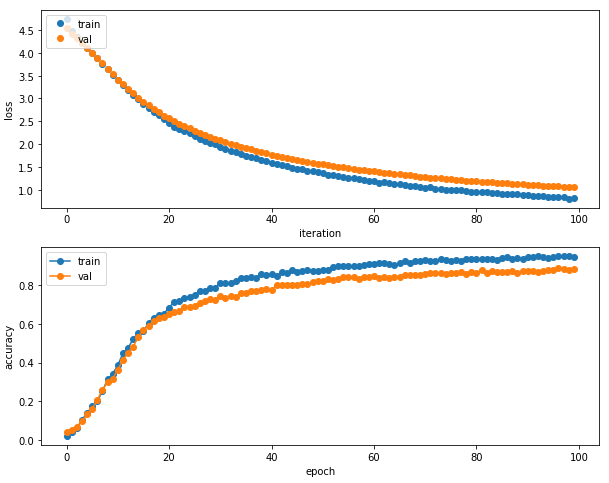}
\includegraphics[width=0.32\linewidth]{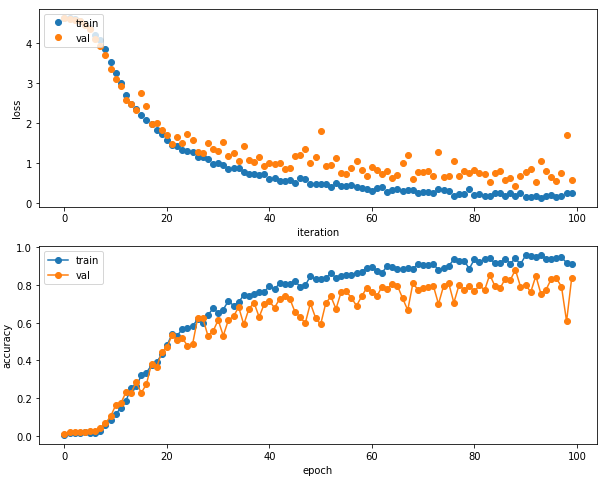}
}
\end{center}
\caption{Loss and Accuracy vs. Epoch \# plots for SqueezeNet 1.1 under Experiments 1 (pre-trained layers finetuned), 2 (pre-trained layers frozen), and 3 (training from scratch, i.e. no pre-training).}\label{apfig:sn}
\end{figure*}

\begin{figure*}[t]
\begin{center}
\fbox{
\includegraphics[width=0.32\linewidth]{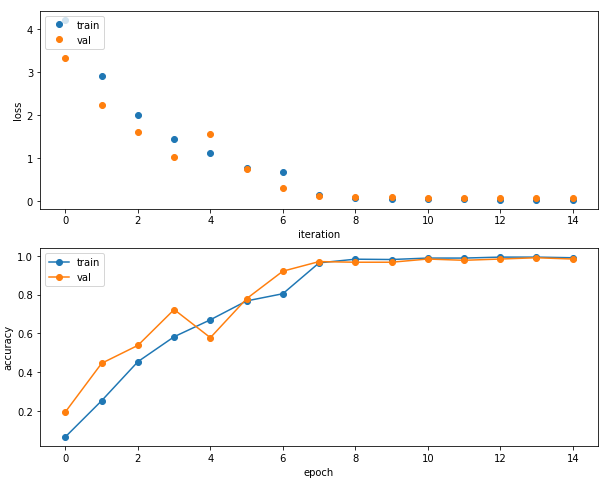}
\includegraphics[width=0.32\linewidth]{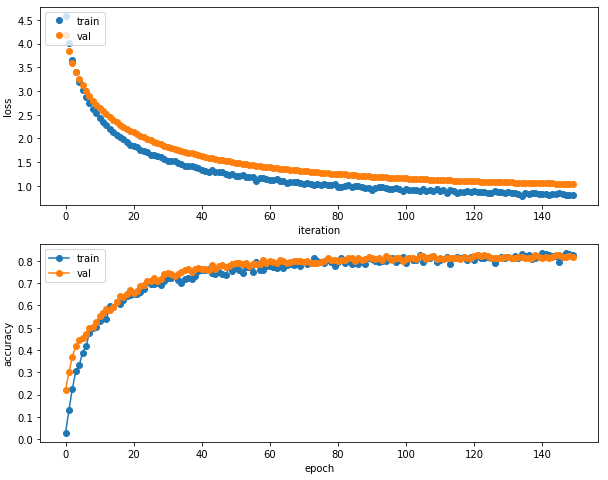}
\includegraphics[width=0.32\linewidth]{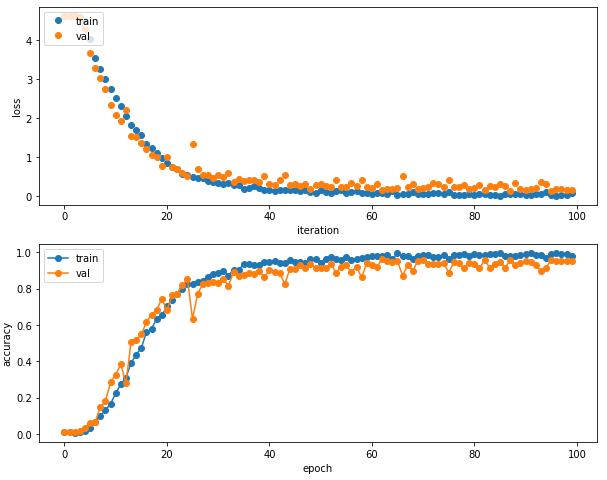}
}
\end{center}
\caption{Loss and Accuracy vs. Epoch \# plots for VGGNet-11 under Experiments 1 (pre-trained layers finetuned), 2 (pre-trained layers frozen), and 3 (training from scratch, i.e. no pre-training).}\label{apfig:vn11}
\end{figure*}

\begin{figure*}[t]
\begin{center}
\fbox{
\includegraphics[width=0.32\linewidth]{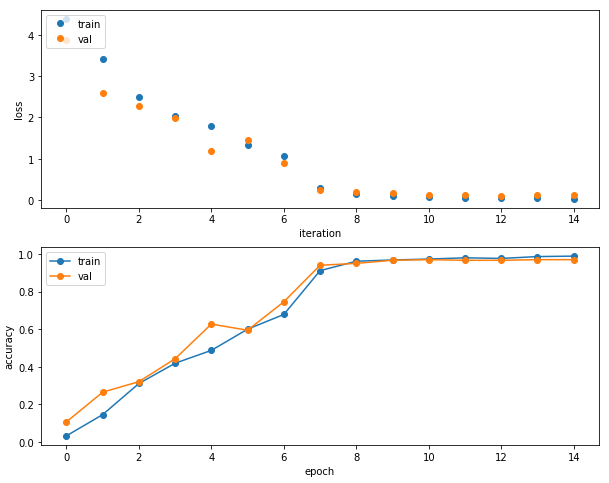}
\includegraphics[width=0.32\linewidth]{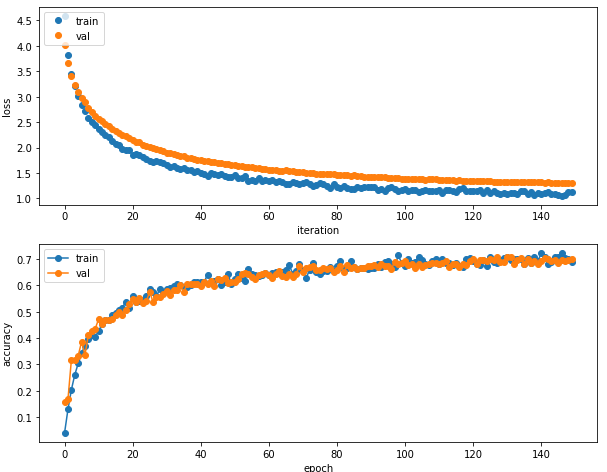}
\includegraphics[width=0.32\linewidth]{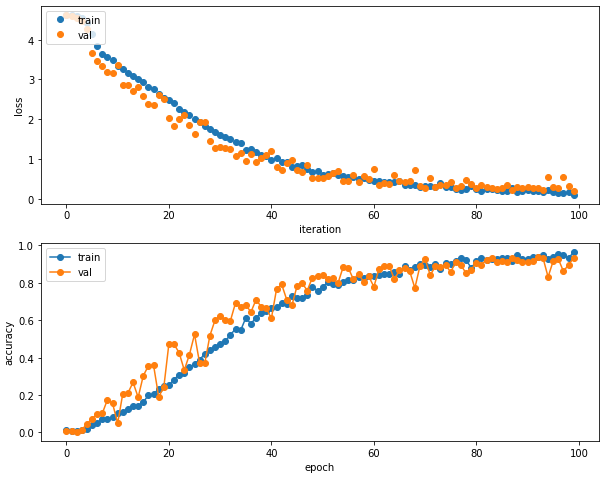}
}
\end{center}
\caption{Loss and Accuracy vs. Epoch \# plots for VGGNet-19 under Experiments 1 (pre-trained layers finetuned), 2 (pre-trained layers frozen), and 3 (training from scratch, i.e. no pre-training).}\label{apfig:vn19}
\end{figure*}

\begin{figure*}[t]
\begin{center}
\fbox{
\includegraphics[width=0.32\linewidth]{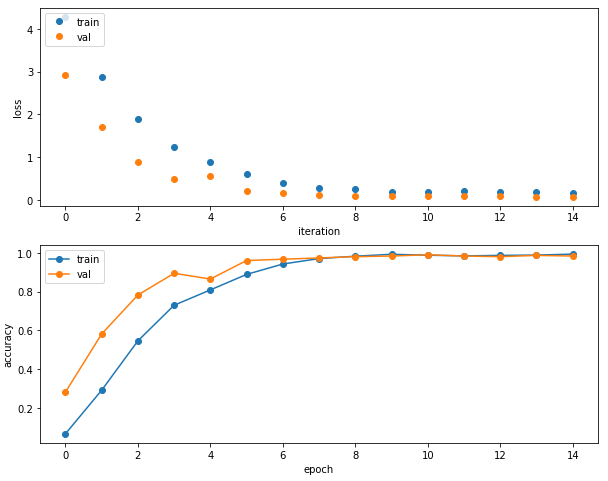}
\includegraphics[width=0.32\linewidth]{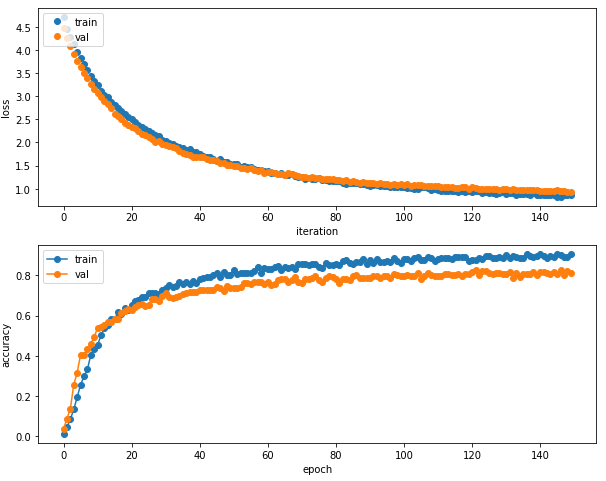}
\includegraphics[width=0.32\linewidth]{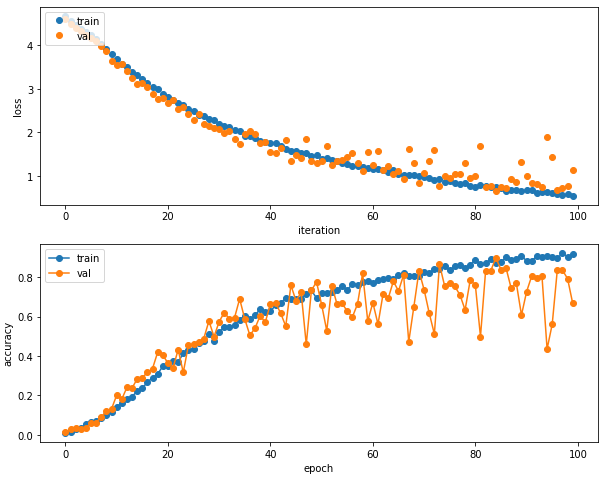}
}
\end{center}
\caption{Loss and Accuracy vs. Epoch \# plots for ResNet-18 under Experiments 1 (pre-trained layers finetuned), 2 (pre-trained layers frozen), and 3 (training from scratch, i.e. no pre-training).}\label{fig:rn18}
\end{figure*}

\begin{figure*}[t]
\begin{center}
\fbox{
\includegraphics[width=0.32\linewidth]{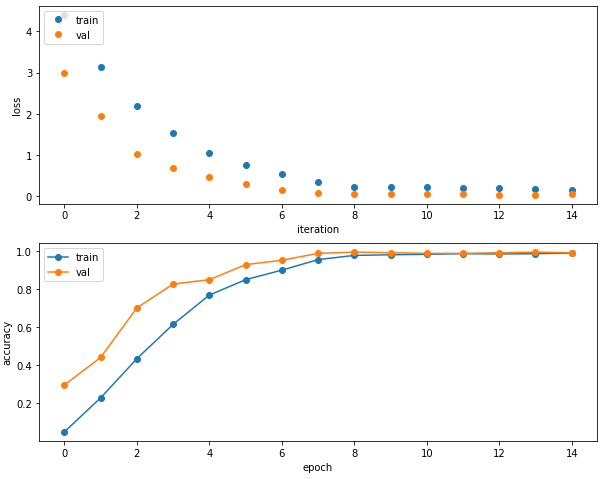}
\includegraphics[width=0.32\linewidth]{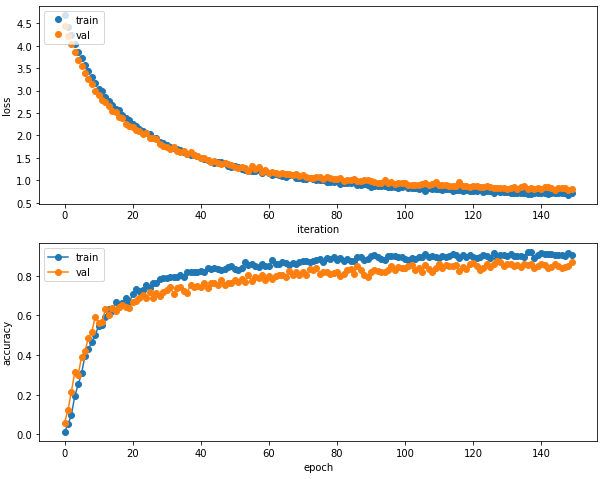}
\includegraphics[width=0.32\linewidth]{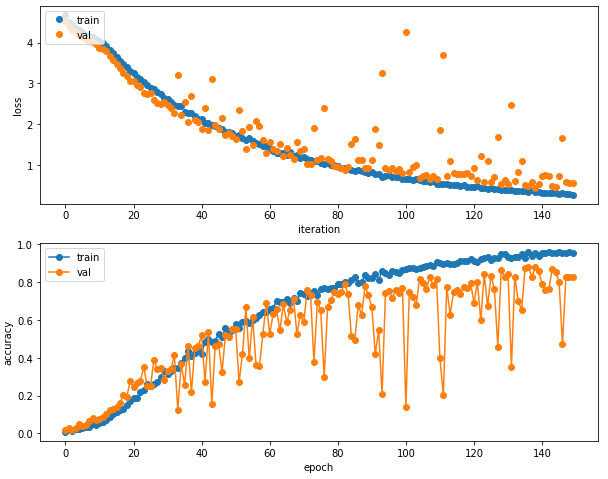}
}
\end{center}
\caption{Loss and Accuracy vs. Epoch \# plots for ResNet-50 under Experiments 1 (pre-trained layers finetuned), 2 (pre-trained layers frozen), and 3 (training from scratch, i.e. no pre-training).}\label{apfig:rn50}
\end{figure*}

Figures~\ref{apfig:astn} through \ref{apfig:rn50} are the complete set of the  accuracy and loss vs. epoch number curves for various network architectures under different training scenarios as discussed in Section~\ref{Exp:MT}.

\vspace{-.5cm}

\section{Contributions \& Acknowledgements}

 The psychological research on human invariance to spatial transformations that motivated this research was  conducted in the Vision \& Perception Neuroscience Lab at Stanofrd University. 
The research used PyTorch software~\cite{PyTorch} and drew upon code from samples in PyTorch tutorials~\cite{STNTutorial,TLTutorial} and PyTorch Vision models~\cite{PyTorchVision} that were pre-trained on ImageNet.

\section{References/Bibliography}
{\small
\bibliographystyle{ieee}
\bibliography{egpaper_final}
}

\end{document}